\documentclass[a4paper,fleqn,colorlinks=true,linkcolor=blue,citecolor=blue,urlcolor=blue]{cas-dc}

\makeatletter\if@twocolumn\PassOptionsToPackage{switch}{lineno}\else\fi\makeatother

\usepackage[authoryear]{natbib} 

\setcitestyle{authoryear,round,semicolon,aysep={,},yysep={,}}

\usepackage{tabulary,xcolor}
\usepackage{amsfonts,amsmath,amssymb}
\usepackage[T1]{fontenc}
\makeatletter
\let\save@ps@pprintTitle\ps@pprintTitle
\def\ps@pprintTitle{\save@ps@pprintTitle\gdef\@oddfoot{\footnotesize\itshape \null\hfill\today}}
\def\hlinewd#1{%
  \noalign{\ifnum0=`}\fi\hrule \@height #1%
  \futurelet\reserved@a\@xhline}

\AtBeginDocument{\ifNAT@numbers \biboptions{sort&compress}\fi}

\makeatother

\def\tsc#1{\csdef{#1}{\textsc{\lowercase{#1}}\xspace}}
\tsc{WGM}
\tsc{QE}

\usepackage{caption}

\DeclareCaptionFont{figfont}{\sffamily\small}

\usepackage{ifluatex}
\ifluatex
\usepackage{fontspec}
\defaultfontfeatures{Ligatures=TeX}
\usepackage[]{unicode-math}
\unimathsetup{math-style=TeX}
\else 
\usepackage[utf8]{inputenc}
\fi 

\usepackage[switch]{lineno}

\usepackage{url,multirow,morefloats,floatflt,cancel,tfrupee}
\makeatletter

\AtBeginDocument{\@ifpackageloaded{textcomp}{}{\usepackage{textcomp}}}
\makeatother
\usepackage{colortbl}
\usepackage{xcolor}
\usepackage{pifont}
\makeatletter

\def\mcWidth#1{\csname TY@F#1\endcsname+\tabcolsep}

\def\cAlignHack{\rightskip\@flushglue\leftskip\@flushglue\parindent\z@\parfillskip\z@skip}
\def\rAlignHack{\rightskip\z@skip\leftskip\@flushglue \parindent\z@\parfillskip\z@skip}

\@ifundefined{etal}{}{}

\usepackage{ifxetex}
\ifxetex\else\if@twocolumn\@ifpackageloaded{stfloats}{}{\usepackage{dblfloatfix}}\fi\fi

\AtBeginDocument{
\expandafter\ifx\csname eqalign\endcsname\relax
\def\eqalign#1{\null\vcenter{\def\\{\cr}\openup\jot\m@th
  \ialign{\strut$\displaystyle{##}$\hfil&$\displaystyle{{}##}$\hfil
      \crcr#1\crcr}}\,}
\fi
}

\AtBeginDocument{%
  \@ifpackageloaded{endfloat}%
   {\renewcommand\efloat@iwrite[1]{\immediate\expandafter\protected@write\csname efloat@post#1\endcsname{}}}{\newif\ifefloat@tables}%
}%

\def\BreakURLText#1{\@tfor\brk@tempa:=#1\do{\brk@tempa\hskip0pt}}
\let\lt=<
\let\gt=>
\def\processVert{\ifmmode|\else\textbar\fi}

\@ifundefined{subparagraph}{
\def\subparagraph{\@startsection{paragraph}{5}{2\parindent}{0ex plus 0.1ex minus 0.1ex}%
{0ex}{\normalfont\small\itshape}}%
}{}

\newcommand\role[1]{\unskip}
\newcommand\aucollab[1]{\unskip}
  
\@ifundefined{tsGraphicsScaleX}{\gdef\tsGraphicsScaleX{1}}{}
\@ifundefined{tsGraphicsScaleY}{\gdef\tsGraphicsScaleY{.9}}{}
\def\checkGraphicsWidth{\ifdim\Gin@nat@width>\linewidth
	\tsGraphicsScaleX\linewidth\else\Gin@nat@width\fi}

\def\checkGraphicsHeight{\ifdim\Gin@nat@height>.9\textheight
	\tsGraphicsScaleY\textheight\else\Gin@nat@height\fi}

\def\fixFloatSize#1{}
\let\ts@includegraphics\includegraphics

\def\inlinegraphic[#1]#2{{\edef\@tempa{#1}\edef\baseline@shift{\ifx\@tempa\@empty0\else#1\fi}\edef\tempZ{\the\numexpr(\numexpr(\baseline@shift*\f@size/100))}\protect\raisebox{\tempZ pt}{\ts@includegraphics{#2}}}}

\AtBeginDocument{\def\includegraphics{\@ifnextchar[{\ts@includegraphics}{\ts@includegraphics[width=\checkGraphicsWidth,height=\checkGraphicsHeight,keepaspectratio]}}}

\DeclareMathAlphabet{\mathpzc}{OT1}{pzc}{m}{it}

\def\URL#1#2{\@ifundefined{href}{#2}{\href{#1}{#2}}}

\def\UrlOrds{\do\*\do\-\do\~\do\'\do\"\do\-}%
\g@addto@macro{\UrlBreaks}{\UrlOrds}

\edef\fntEncoding{\f@encoding}

\makeatother

\newif\ifmultipleabstract\multipleabstractfalse%
\usepackage{float}

\usepackage{hyperref}

\begin{document}

\let\WriteBookmarks\relax
\def\floatpagepagefraction{1}
\def\textpagefraction{.001}

\shorttitle{3D Characterization of Smoke Plume Dispersion Using Multi-View Drone Swarm}

\shortauthors{Nikil Krishnakumar \MakeLowercase{\textit{et al.}} }
    \title[mode = title]{    
  3D Characterization of Smoke Plume Dispersion Using Multi-View Drone Swarm    
}
    
\author[a3aaeda92b17e]{Nikil Krishnakumar}

\author[a3aaeda92b17e]{Shashank Sharma}

\author[a3aaeda92b17e]{Srijan Kumar Pal}

\author[ad1ee938236e5]{Jiarong Hong \fnref{fn1}}\cormark[1]
\ead{jhong@umn.edu}

\cortext[1]{Corresponding author.}
\fntext[fn1]{\href{https://orcid.org/0000-0001-7860-2181}{ORCID: 0000-0001-7860-2181}}




\address[a3aaeda92b17e]{Minnesota Robotics Institute\unskip, 
    St. Anthony Falls Laboratory\unskip, University of Minnesota\unskip, Minneapolis\unskip, Minnesota\unskip, United States}
  	
\address[ad1ee938236e5]{Mechanical Engineering \unskip, 
    St. Anthony Falls Laboratory\unskip, University of Minnesota\unskip, Minneapolis\unskip, Minnesota\unskip, United States}

\begin{abstract}
This study presents an advanced multi-view drone swarm imaging system for the three-dimensional characterization of smoke plume dispersion dynamics. The system comprises a manager drone and four worker drones, each equipped with high-resolution cameras and precise GPS modules. The manager drone uses image feedback to autonomously detect and position itself above the plume, then commands the worker drones to orbit the area in a synchronized circular flight pattern, capturing multi-angle images. The camera poses of these images are first estimated, then the images are grouped in batches and processed using Neural Radiance Fields (NeRF) to generate high-resolution 3D reconstructions of plume dynamics over time. Field tests demonstrated the system’s ability to capture critical plume characteristics including volume dynamics, wind-driven directional shifts, and lofting behavior at a temporal resolution of about 1 s. The 3D reconstructions generated by this system provide unique field data for enhancing the predictive models of smoke plume dispersion and fire spread. Broadly, the drone swarm system offer a versatile platform for high resolution measurements of pollutant emissions and transport in wildfires, volcanic eruptions, prescribed burns, and industrial processes, ultimately supporting more effective fire control decisions and mitigating wildfire risks.
\end{abstract}
    
\maketitle
\makeatletter
\def\ps@pprintTitle{\let\@oddfoot\@empty\let\@evenfoot\@empty}
\makeatother

\section{Introduction}
Understanding the transport dynamics of atmospheric particles, such as dust, snow, smoke, and sand, is essential due to its significant impact on air quality, climate, and ecological systems across various environmental processes, including wildfires, sandstorms, snowstorms, and volcanic eruptions \unskip~\citep{mott2010understanding, kumar2011dynamics, kok2012physics, butwin2019effects, evangeliou2020atmospheric, jaffe2020wildfire, dentoni2022emission}. This is important for prescribed burns, which are controlled fires used in forest management to enhance ecological health and reduce the risk of wildfires. However, the occurrence of 43 wildfires out of 50,000 prescribed burns in the U.S. between 2012 and 2021 underscores the need for effective smoke management to ensure public safety and minimize adverse effects on air quality \unskip~\citep{AssociatedPress2024}. The challenges in managing these burns highlight a critical gap in our understanding of the dispersion dynamics of particles during these events, which can lead to severe and potentially devastating outcomes \unskip~\citep{kalabokidis2000effects,pereira2021short}. Therefore, there is a pressing need for comprehensive research to better predict, manage, and mitigate the risks associated with prescribed burns.

To address these challenges, researchers are developing various simulation tools, such as QUIC-Fire and FIRETEC \unskip~\citep{linn2002studying, linn2020quic}, that aim to model fire and smoke particle behavior. These tools utilize complex inputs, including 3D maps of fuel sources, vegetation structure, topography, moisture content, and wind predictions \unskip~\citep{rowell2020coupling, robinson2023quic}. Despite these advancements, significant limitations remain. There is a lack of validation that compares the predicted movement of particles with actual plume dispersion, a shortage of dynamic 3D ground truth data on particle dispersion, and difficulties in making accurate predictions in areas where 3D fuel data is unavailable \unskip~\citep{linn2020quic,Sara2023}. These limitations underscore the ongoing need for field data to validate and improve these models, ensuring more accurate predictions and better management.

However, current field tools for data collection have significant limitations. Remote sensing and Lidar technologies, while valuable, lack the spatial and temporal resolution required to capture the highly dynamic flows of smoke plumes during prescribed burns \unskip~\citep{prichard2019fire, sokolik2019progress}. Moreover, these tools are constrained by their limited mobility, making it difficult to effectively monitor events across varied terrains or in remote and inaccessible areas. The inability to collect such detailed data hinders the development of more accurate and reliable models for managing prescribed burns and predicting smoke dispersion.

To address these challenges, this study aims to develop an autonomous drone swarm equipped with cameras to capture multi-angle images of smoke plumes. This approach will enable the 3D reconstruction of plume dispersal dynamics, allowing for detailed analysis of flow patterns. By deploying a fleet of drones for multi-view imaging, we intend to create a comprehensive 3D ground truth model of specific burn events. This model will provide researchers with critical data for validating simulation predictions and offer essential guidance for hazard response and management.
While there has been no prior work on 3D reconstruction of particle transport using multi-view images from drones, significant advancements have been made in 3D reconstruction techniques with static objects 
\unskip~\citep{schenk2005introduction,goesele2006multi,schonberger2016structure,hackl2018use,mildenhall2021nerf,muller2022instant}. Prominent methods include Structure from Motion (SfM) and Multi-view Stereo (MVS), which reconstructs 3D models from 2D image sequences through feature tracking and photogrammetry, allowing for precise estimation of camera poses and 3D structures \unskip~\citep{goesele2006multi,schonberger2016structure}. Neural Radiance Fields (NeRF) have further pushed the boundaries of scene reconstruction by generating photorealistic views through the optimization of a continuous 5D neural radiance field from sparse input images \unskip~\citep{mildenhall2021nerf,muller2022instant}. Building on these advancements, D-NeRF extends this capability to dynamic scenes, capturing non-rigid motion and deformation over time \unskip~\citep{pumarola2021d}. Additionally, RoDynRF enables dynamic view synthesis from monocular videos, even without known camera poses \unskip~\citep{liu2023robust}. However, reconstructing atmospheric dispersion plumes presents unique challenges. SfM struggles with the featureless nature of plumes \unskip~\citep{schonberger2016structure}, NeRF is primarily designed for static scenes \unskip~\citep{mildenhall2021nerf,muller2022instant}, D-NeRF may not perform well with unfamiliar or highly variable scenes \unskip~\citep{pumarola2021d}, and RoDynRF faces difficulties with the complex dynamics of plumes and demands extensive training times \unskip~\citep{liu2023robust}. These limitations can be effectively addressed using our multi-view drone swarm approach. By deploying multiple drones, we can adaptively capture images of the plume at various scales and positions, tailored to the plume's size and dynamic evolution. With this adaptive imaging strategy, we can fully harness the efficiency (compared to D-NeRF) and accuracy of the NeRF pipeline, generating highly detailed 3D reconstructions for each temporal snapshot. This approach allows us to capture multiple reconstructions over time, facilitating the study of the plume's dispersion change.

The structure of this paper is as follows: Section II details the proposed drone swarm platform for 3D plume characterization, including both the drone hardware and 3D reconstruction method. Section III demonstrates the efficiency of our system through field deployment of our multi-view drone swarm and the follow-up 3D plume reconstruction and plume characterization. Finally, we summarize our findings and discuss their implications and limitations.

\section{Methodology}

\subsection{ Overview}
\bgroup
\begin{figure}[!htbp]
\centering
\includegraphics{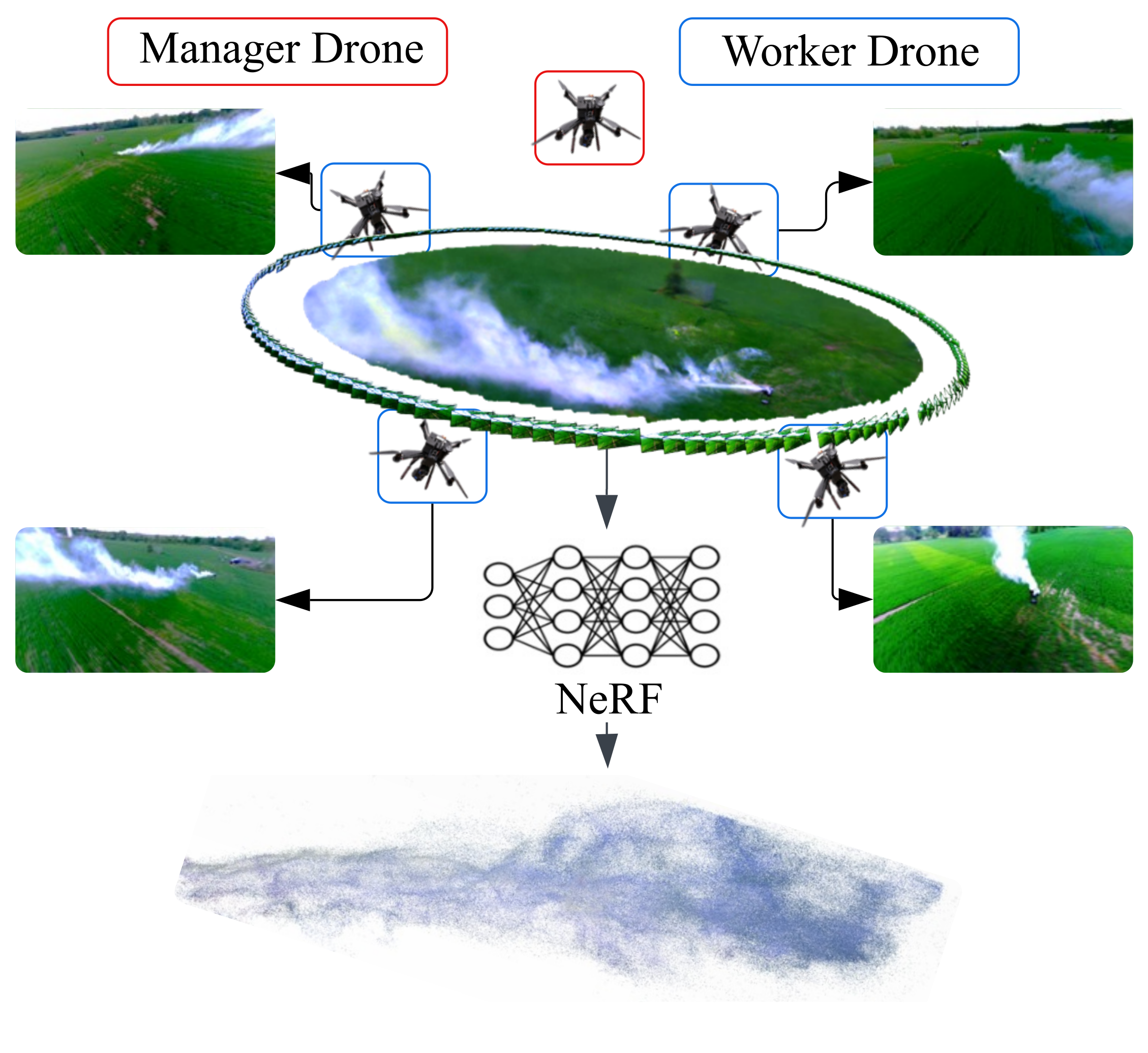}
\caption{Illustration of the drone swarm system that uses multi-view imaging for 3D smoke plume characterization.}
\label{figure-1}
\end{figure}

\egroup
As illustrated in \textcolor{blue}{Figure~\ref{figure-1}} , our drone swarm system for multi-view imaging comprises two main components: the data acquisition module and the data processing module. The data acquisition module includes a manager drone and four worker drones, which work in a coordinated manner to capture multi-view images of the smoke plume and its evolution over time. These drones are equipped with synchronized imaging systems that allow them to document the plume at different time steps from various angles.

In the data processing module, the captured images from the worker drones are compiled and segregated by segmented time intervals. The images are then fed into the NeRF network, and the output point cloud from this is further processed to remove the background and segment the plume in 3D. This 3D model is used to extract important characteristics of the plume, such as its volume, angle of deviation, and other dynamics of plume dispersion in the atmosphere. The process is repeated for each time segment to provide a comprehensive spatial and temporal characterization of the plume dynamics.

\subsection{ Data Acquisition Module}
\bgroup
\begin{figure}[h!]
\centering
\includegraphics{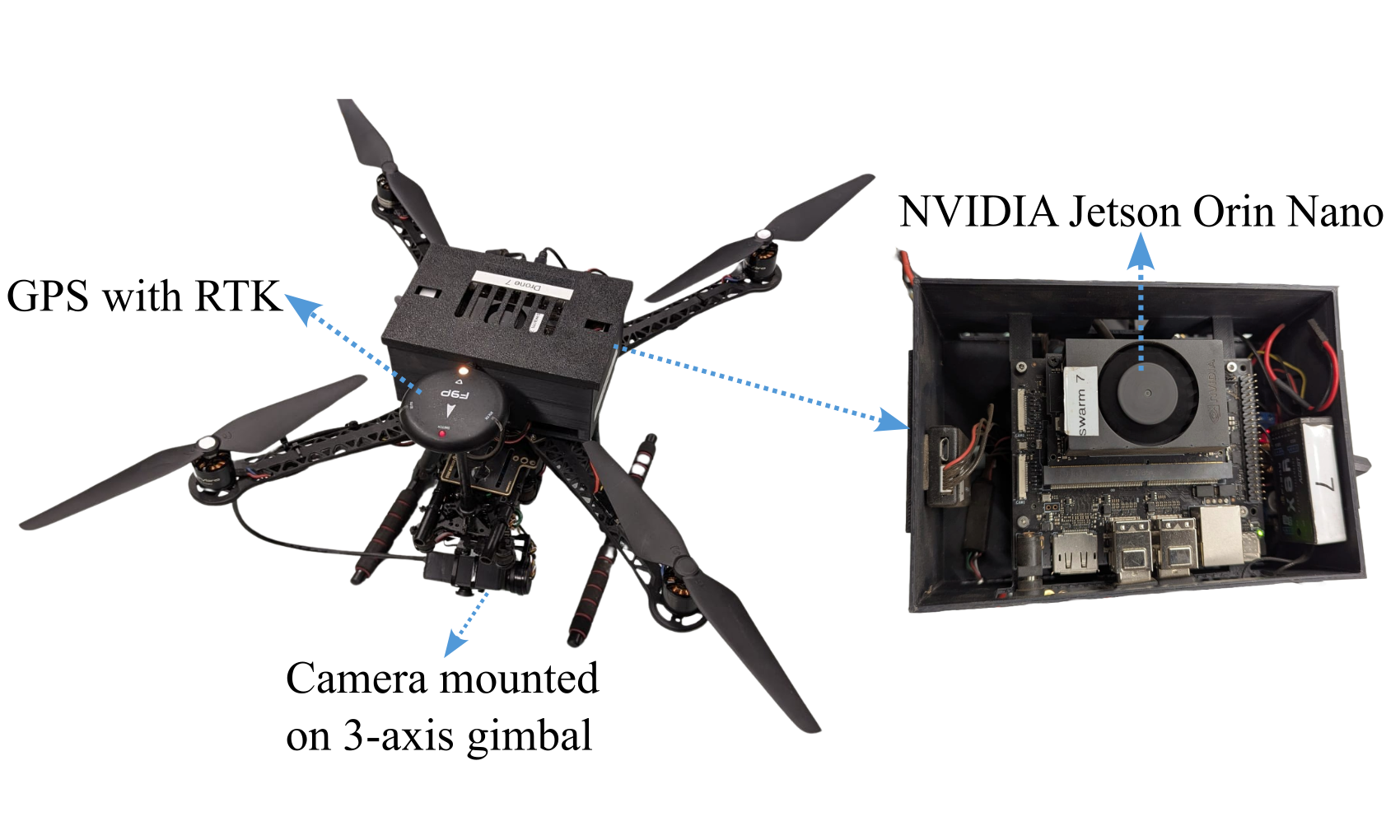}
\caption{Drone hardware configuration showing the quadcopter with camera mounted on a 3-axis gimbal and GPS with RTK (left), and the NVIDIA Jetson Orin Nano (right).}
\label{figure-2}
\end{figure}
\egroup
\bgroup
\begin{figure*}[!htbp]
\centering
\includegraphics{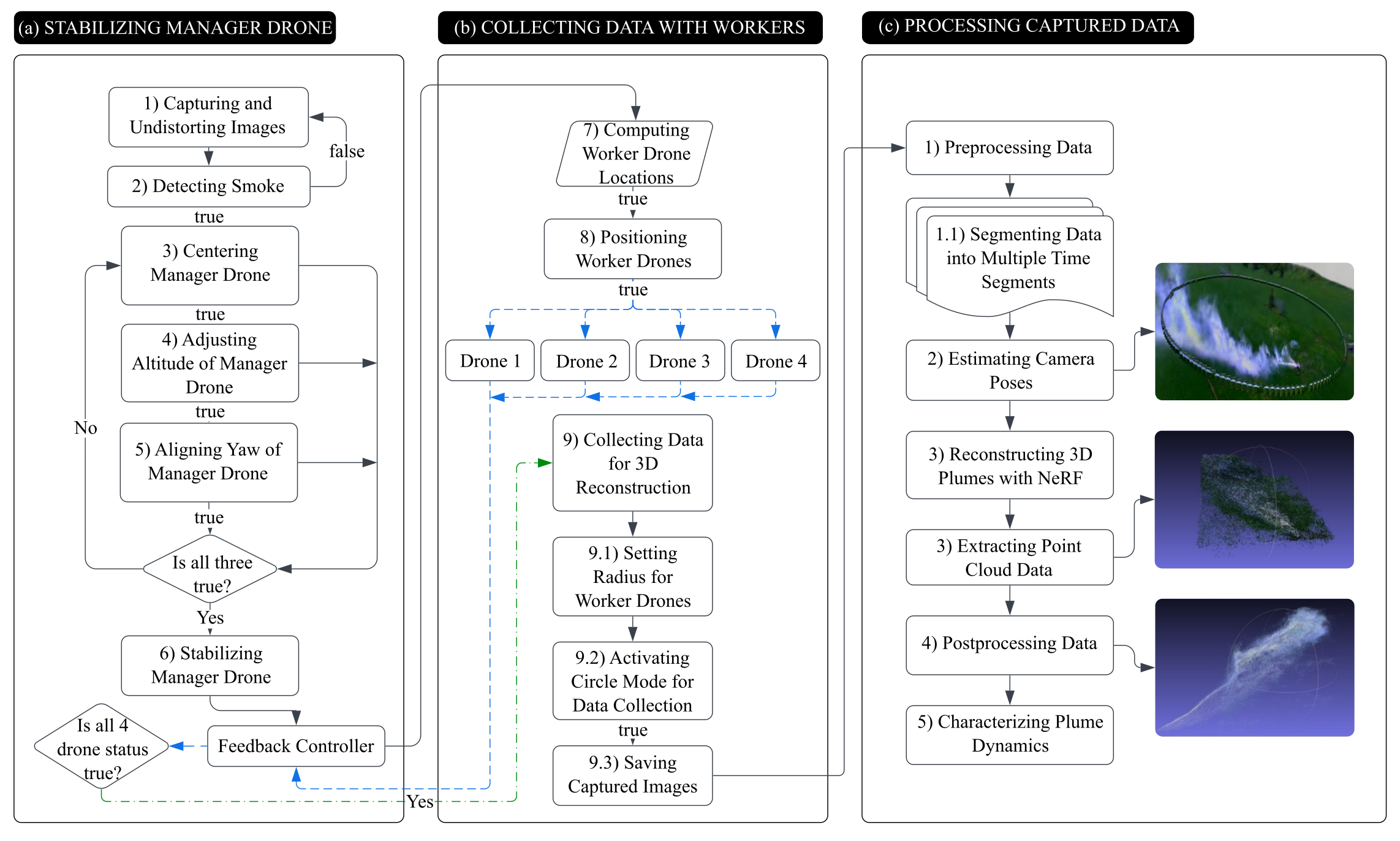}
\caption{Flowcharts detailing the steps involved in (a) stabilizing the manager drone, (b) collecting data with the worker drone swarm, and (c) processing captured data for 3D plume reconstruction and characterization.}
\label{figure-3}
\end{figure*}
\egroup
The drone swarm system consists of a manager drone and four worker drones, built on durable Holybro S500 V2 quadcopter frames. Each drone is equipped with a 12 Megapixel (MP) ArduCam USB-based camera mounted on a 3-axis gimbal for smoke detection, image feedback control, and dataset collection to facilitate 3D reconstruction, as shown in \textcolor{blue}{Figure~\ref{figure-2}}. The drones are powered by 6000 mAh lithium-polymer batteries and controlled using Holybro Pixhawk 6C flight controllers running ArduPilot. Commands can be transmitted through a 2.4 GHz FrSky RC controller, a 915 MHz telemetry radio via MAVLink, or directly over USB using the NVIDIA Jetson platform. These components are similar to those used in \citep{bristow2023atmospheric}.

The manager drone is equipped with an NVIDIA Jetson Orin Nano, while the worker drones use the NVIDIA Jetson Nano, with the manager drone having a more powerful computer to handle the computational demands of smoke detection and navigation control for both itself and the worker drones. All drones run MAVROS, which facilitates communication between the onboard computer and the flight controller by creating ROS topics. These topics consist of subscribers that receive sensor data such as GPS location, altitude, orientation, velocity, timestamps, and drone state, while publishers send control commands to the flight controller for autonomous operations. The manager drone operates on ROS Noetic, while the worker drones run ROS Melodic, ensuring compatibility with their respective hardware. MAVROS also enables WiFi-based broadcasting of all ROS topics, allowing each drone to access real-time data from others, supporting inter-sensor communication and drone-to-drone coordination for autonomous swarm operations.

To enable coordinated data collection, the drones are interconnected via a robust outdoor Wi-Fi network with a speed of approximately 1775 Mbps, providing stable wireless coverage up to 200-300 m at the 5 GHz band. Additionally, for high-precision positioning, we employ Real-Time Kinematic (RTK) technology, which provides centimeter-level accuracy by utilizing carrier phase measurements from GNSS signals, achieved through triangulation between the RTK base station, GPS, and satellites. The drones operate based on a swarm control algorithm, depicted in \textcolor{blue}{Figure~\ref{figure-3}}, maintaining optimal spacing and coverage around the plume to capture images from multiple perspectives.

1) Capturing and Undistorting Images: To ensure accurate 3D reconstruction and image feedback control, camera calibration is performed to determine the camera matrix and distortion coefficients, which are then used to undistort images. In this setup, the manager drone captures images at a resolution of 640 x 480 pixels and processes (segmentation) at a rate of ten fps for realtime feedback, while worker drones use a resolution of 1280 x 720 pixels for better image reconstruction. The full-sensor size of the camera is not used due to its low frame rate and the excessive computational load it would impose on the Jetson Nano.

2) Detecting Smoke: The smoke plume detection is performed using a YOLOv8-Segmentation model \citep{Jocher2023}, specifically the YOLO-Seg-N variant, which provides the lightest model weights to maximize inference speed. A dataset of top-down smoke plume images is collected using a drone under varying background and lighting conditions. The dataset is constructed by manually annotating smoke regions, followed by a train-validation-test split of 66.7\%–16.7\%–16.7\%, ensuring no overlap between subsets. The model undergoes 400 training epochs, allowing it to generalize effectively to unseen smoke plume scenarios. Field testing demonstrates effective plume detection, enabling autonomous operations in real-world environments.

3) Centering Manager Drone: The drone tracks and centers on the plume's centroid obtained from plume segmentation in the previous step by adjusting its position until the centroid is within a specified threshold of the image center. It accomplishes this by publishing velocity commands (cmd\_vel) with twist values for the x and y axes to the drone via MAVROS. These commands move the drone in the direction of the centroid. The process continues until the centroid is aligned with the image center, ensuring accurate tracking.

4) Adjusting Altitude of Manager Drone: The drone adjusts its altitude based on the segmented smoke area, ensuring optimal positioning for effective tracking. If the smoke area exceeds an upper threshold, the drone ascends; conversely, if the smoke area falls below a lower threshold, the drone descends. This process continues until the drone reaches the optimal range, where the smoke area comprises 8\% to 12\% of the image. This threshold is chosen to ensure that most of the smoke is captured within the frame, while still including a portion of the background. Maintaining this balance helps in accurately positioning the drone for swarm operations, allowing the drones to surround the smoke plume effectively and coordinate the mission.

5) Aligning Yaw of Manager Drone: To align the drone perpendicularly to the flow of the dispersion plume, the covariance of the segmented mask is calculated, resulting in eigenvectors that indicate the plume's flow direction relative to the image. The drone is then yawed perpendicular to the largest eigenvector. Using the drone's current heading, yaw adjustment is calculated and executed to achieve desired orientation.

6) Stabilizing Manager Drone: The three processes are re-initiated repeatedly until all parameters fall within their thresholds. Once these conditions are met, the drone stabilizes and maintains its position, ensuring it is correctly aligned and in the optimal location.

7) Computing Worker Drone Locations: Using the drone's GPS coordinates, altitude, and camera focal length, we calculate the real-world dimensions of the captured image. By applying the haversine formula, we calculate the latitude and longitude of the image's corners, based on the known latitude and longitude of the image center. Using this information, we compute an affine transformation matrix with the least error, that maps each pixel to its corresponding GPS coordinates.

8) Positioning Worker Drones: To precisely localize and position the drones around the plume from four sides, the target locations for each drone are calculated based on data from the manager drone. The manager drone captures a 640 x 480-pixel image, and the two vertical extreme points of this image are identified to maximize coverage of the smoke plume while collecting data. The distance from the center to these vertical extremes is used to determine corresponding horizontal positions, ensuring that all worker drones are equidistant from the center. The positions will adjust with each run based on the size of the plume, as the altitude of the manager drone changes accordingly to accommodate plume size variations. Once these positions are computed, the drones are dispersed to their designated locations at the desired altitude, yawing to face the manager drone for optimal data collection.

9) Collecting Data for 3D Reconstruction: Worker drones maintain their positions until they reach their designated locations as assigned by the manager drone. Once in position, each worker drone verifies that its current GPS location is within a safe offset of 0.5 m from the assigned coordinates. Upon confirmation, the drone transmits a readiness signal to the manager. When all drones have reported readiness, the manager drone issues a simultaneous command for the worker drones to establish the desired circular formation, with the radius determined by the distance between the manager and the worker drones. This configuration ensures that all drones remain equidistant from the manager drone and effectively encompass the targeted plume, optimizing spatial resolution for 3D plume reconstruction. Once the formation is set, the drones transition to circling mode, maintaining their designated orbit while capturing plume images. This synchronized movement ensures precise data acquisition and consistent image quality across the swarm, thereby enhancing the accuracy and reliability of the 3D reconstruction process.

\subsection{ Data Processing Module}The data processing module consists of several stages designed to efficiently handle the data captured by the drones, as depicted in the flow chart in \textcolor{blue}{Figure~\ref{figure-3}c}.

1) Preprocessing Data: Each drone captures data corresponding to one-quarter of a circular region. When combined, data from all four drones form a complete circle around the plume, representing a single time segment. As the drones continue capturing data, additional circles are generated, each corresponding to a new time segment. To enhance temporal resolution, data overlaps are introduced between two consecutive time segments, effectively creating additional intermediate time segments

2) Estimating Camera Poses: Distinctive images captured by each drone are labeled and the aggregation of this is fed into COLMAP, which employs SIFT feature extraction, exhaustive feature matching, structure-from-motion, and bundle adjustment. This process estimates camera poses for all drones relative to one another, ensuring that independent reconstructions are aligned within a unified coordinate system.

3) Reconstructing 3D Plumes with NeRF: For each time segment, the required camera poses are extracted from the total poses computed by COLMAP. Using the COLMAP camera trajectory and corresponding image data, a NeRF model is trained to reconstruct the 3D structure of the plume\unskip~\citep{mildenhall2021nerf}. During this process, 2D image data is projected into radiance fields, which encode spatial density and view-dependent color variations for each point in the scene. NeRF learns a continuous volumetric representation of the plume by optimizing a fully connected neural network, mapping input camera rays to density and color values. This enables NeRF to synthesize novel views and interpolate missing visual data between captured angles, producing a more complete and smooth 3D representation of the smoke plume. To refine the output, the resulting 3D data outside the defined enclosure region is cropped, ensuring that only the relevant smoke volume is retained. The final reconstructed plume structure is then exported as point clouds

4) Post-Processing Data: Background removal is performed using a combination of YOLO-v8 and a Naive Bayes Gaussian model. YOLO-v8 is used to detect and segment smoke plumes in three randomly selected images from the input data. These segmented masks, along with the background RGB data, are then used to train a Naive Bayes Gaussian classifier. The classifier is employed to segment the smoke plumes from the point clouds, effectively removing the background.

5) Characterizing Plume Dynamics: In the final step, the processed 3D model is analyzed to extract crucial plume characteristics, such as spatial extent and expansion angle. These features are essential for understanding plume dynamics and supporting the development of predictive models for behavior in various atmospheric conditions.
    
\section{System Demonstration}
\subsection{Field Deployment}

\bgroup
\begin{figure}[h]
\centering
\includegraphics{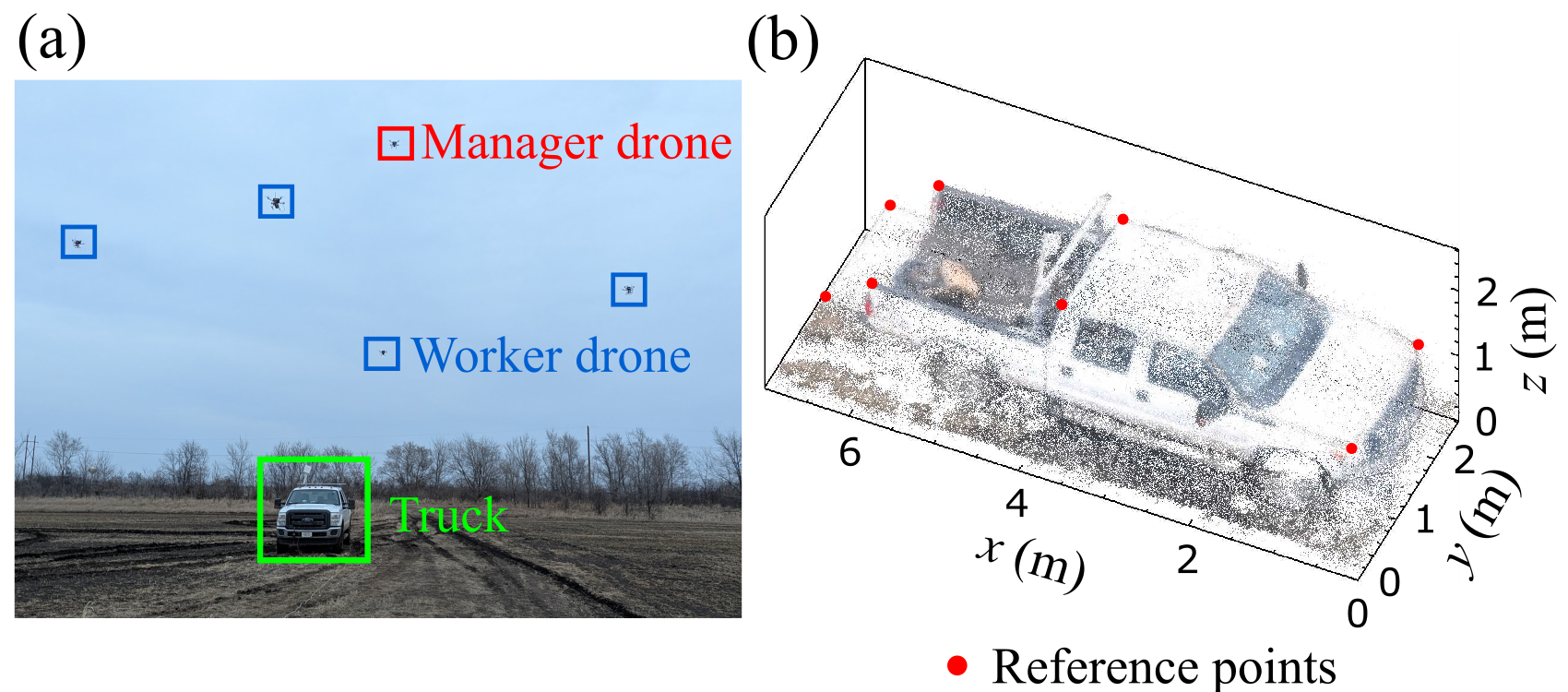}
\caption{3D Reconstruction of truck to validate accuracy (a) Field setup and (b) the corresponding 3D reconstructed point cloud of a 2011 Ford F-350 pickup truck, generated using our multi-view drone swarm imaging system to evaluate its 3D reconstruction accuracy.}
\label{figure-4}
\end{figure}
\egroup
We conducted field testing in two phases. The first phase focused on assessing our drone system’s 3D reconstruction accuracy using a static reference object—specifically a 2011 Ford F‑350 pickup truck—while the second phase demonstrated its ability to capture the 3D dispersion of a smoke plume. Because smoke is inherently dynamic, validating reconstruction fidelity under real plume conditions is challenging. Thus, in the first phase as shown in \textcolor{blue}{Figure~\ref{figure-4}}, we deployed the drones at an altitude of approximately 10 m in a circular trajectory around the truck (20 m in diameter), capturing images from multiple perspectives. After generating a point cloud from these images, we scaled it and measured reconstruction accuracy against eight known reference points on the truck. The results showed an average error of roughly 1.18 \% with a standard deviation near 0.98 \%, confirming the system’s capacity to accurately capture 3D structures and establishing a foundation for the second phase’s focus on dynamic smoke plumes.

\bgroup
\begin{figure}[h]
\centering
\includegraphics{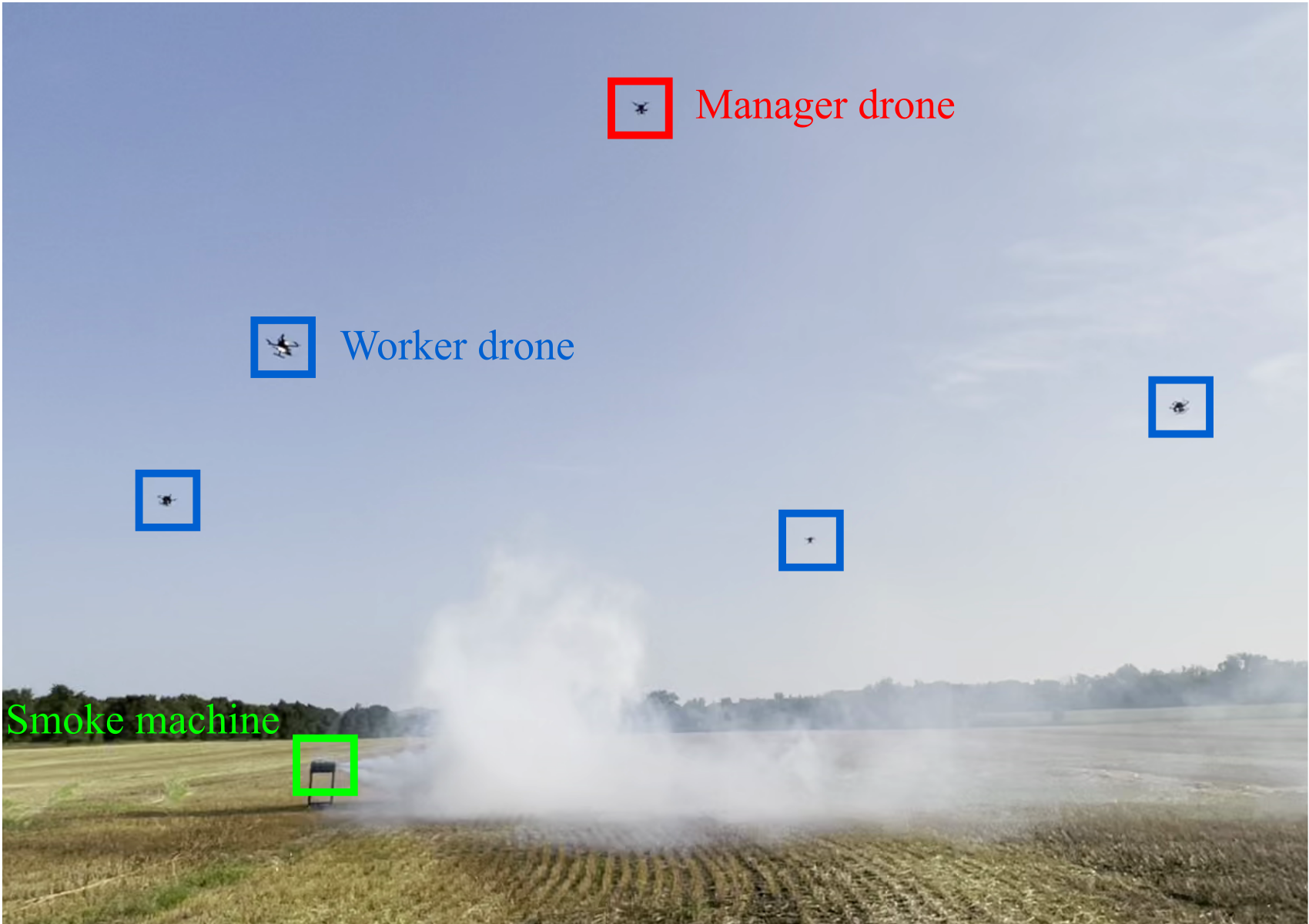}
\caption{Field deployment setup for data collection, featuring a manager drone positioned above the plume for centralized control and four worker drones encircling the plume to capture multi-angle images for 3D reconstruction.}
\label{figure-5}
\end{figure}
\egroup

Following the static validation, we conducted the second major experiment to evaluate the system’s performance in reconstructing dynamic smoke plumes, as shown in \textcolor{blue}{Figure~\ref{figure-5}}. For this test, smoke plumes were generated using a high-density smoke generator, which utilizes a non-harmful smoke fluid composed of high-density fog liquid, food-grade glycerine, and propylene glycol. The generated smoke typically extended up to 40 m in length, with variability depending on the smoke machine's emission intensity. To enhance plume production and density, we employed two smoke machines—one producing a high volume of smoke that diminished and regenerated cyclically, while the other operated intermittently to optimize overall density. Together, these machines generated plumes with widths ranging from 1 to 10 m and a maximum height of 10 m.

Once the smoke generator and Wi-Fi network were set up, the drones were powered on and connected to the network. We initiate MAVROS nodes in each drone via Secure Socket Shell (SSH), with the manager drone serving as the ROS Master. From the base station, commands were executed to begin operations. The manager drone was launched manually first to process images and relay data to the base station. Upon 
detecting smoke, the drone was switched to GUIDED mode to autonomously position itself above the plume. Following this, worker drones were launched and set to GUIDED mode to autonomously adjust their positions and optimize coverage based on plume size.

In the experiment, the drones followed circular paths with an average radius of 21 m around the plume. Each drone completed a full circle in approximately 32 s, recording data at eight fps. Equipped with 6000 mAh, 4S batteries, each drone could perform up to five complete circles before experiencing performance degradation, such as altitude drops due to reduced thrust voltage. At this proof-of-concept stage, the system provided two minutes and 20 s of stable flight time, allowing for five full data collection circuits. All data were recorded onboard for post-flight analysis.

\subsection{ 3D Plume Reconstruction}
\bgroup
\begin{figure*}[b!]
\centering
\includegraphics{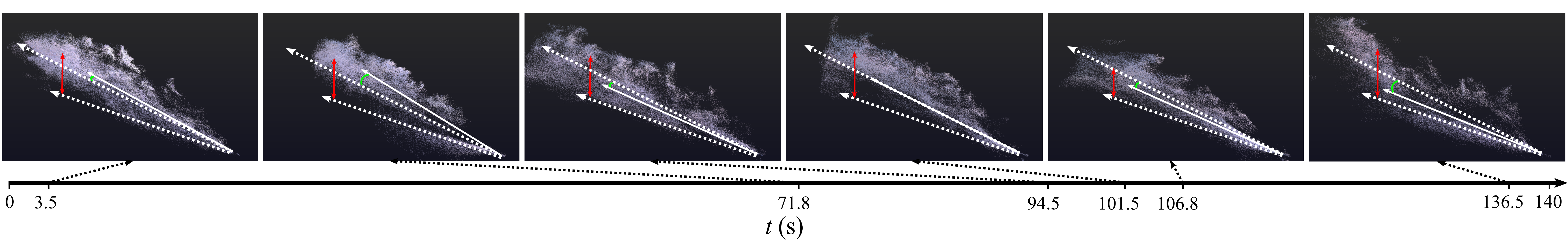}
\caption{Snapshots of the 3D reconstructed plume showing variations in volume, direction, and shape over the 140 s sampling period. Major dotted lines indicate reference lines for the angle of deviation, with the green-highlighted angle between the major dotted line and the white solid line representing the angle of deviation. Minor dotted lines serve as reference lines for average height, while the vertical red line highlights the average height.}
\label{figure-6}
\end{figure*}
\egroup

In this study, we applied NeRF to reconstruct the 3D dynamics of a smoke plume over two minutes and 20 s recording interval, during which each drone completed five revolutions around the plume. The reconstruction was based on images captured by four drones, each circling the plume at quarter-circle intervals. Each drone required eight seconds to complete a quarter-circle, and because all drones operated synchronously, the combined data provided a full-circle dataset every eight seconds. During each quarter-circle, a drone captured 65 images, resulting in a total of 260 images per full revolution. These images were then processed to reconstruct the plume in 20 distinct time segments, with each segment covering approximately eight seconds of plume dynamics.

To enhance temporal resolution and capture smoother plume dynamics, we introduced overlaps of 25\%, 50\%, and 75\% between time segments, generating three additional reconstructions between each pair of segments. This approach resulted in a total of 77 reconstructed time segments, providing a finer temporal resolution of 1.75 s. This adjustment enabled a more detailed and continuous observation of plume behavior over time.

The 3D reconstruction was performed using a high-performance computing system equipped with a 13th Gen Intel Core i7-13700K CPU, 64 GB of RAM, and an NVIDIA RTX 5000 Ada GPU with 32 GB of memory. The computational time for each 8-second segment was approximately 10 minutes. The process for reconstructing and saving filtered point clouds from segregated data has been fully automated, ensuring efficiency and consistency in data handling.

As shown in \textcolor{blue}{Figure~\ref{figure-6}}, the reconstructed models capture significant changes in plume dynamics over the recording period. Snapshots reveal the plume’s variations in volume, direction, and shape. Early in the sampling period, the plume exhibits an average volume and elevation, while later stages show greater dispersion, distinct directional deviations, and reduced density. Key stages include initial growth and ascent, lateral dispersion under wind influence, and eventual dissipation with diminished volume and height. The reconstructed models reveal critical changes in the plume’s evolution over time, showcasing its growth, directional shifts, and eventual dissipation. These dynamic reconstructions lay the groundwork for quantitative analysis of plume characteristics, discussed in the following section.

\subsection{Quantitative Characterization of Plume Dynamics}
\bgroup
\begin{figure}[h]
\centering
\includegraphics{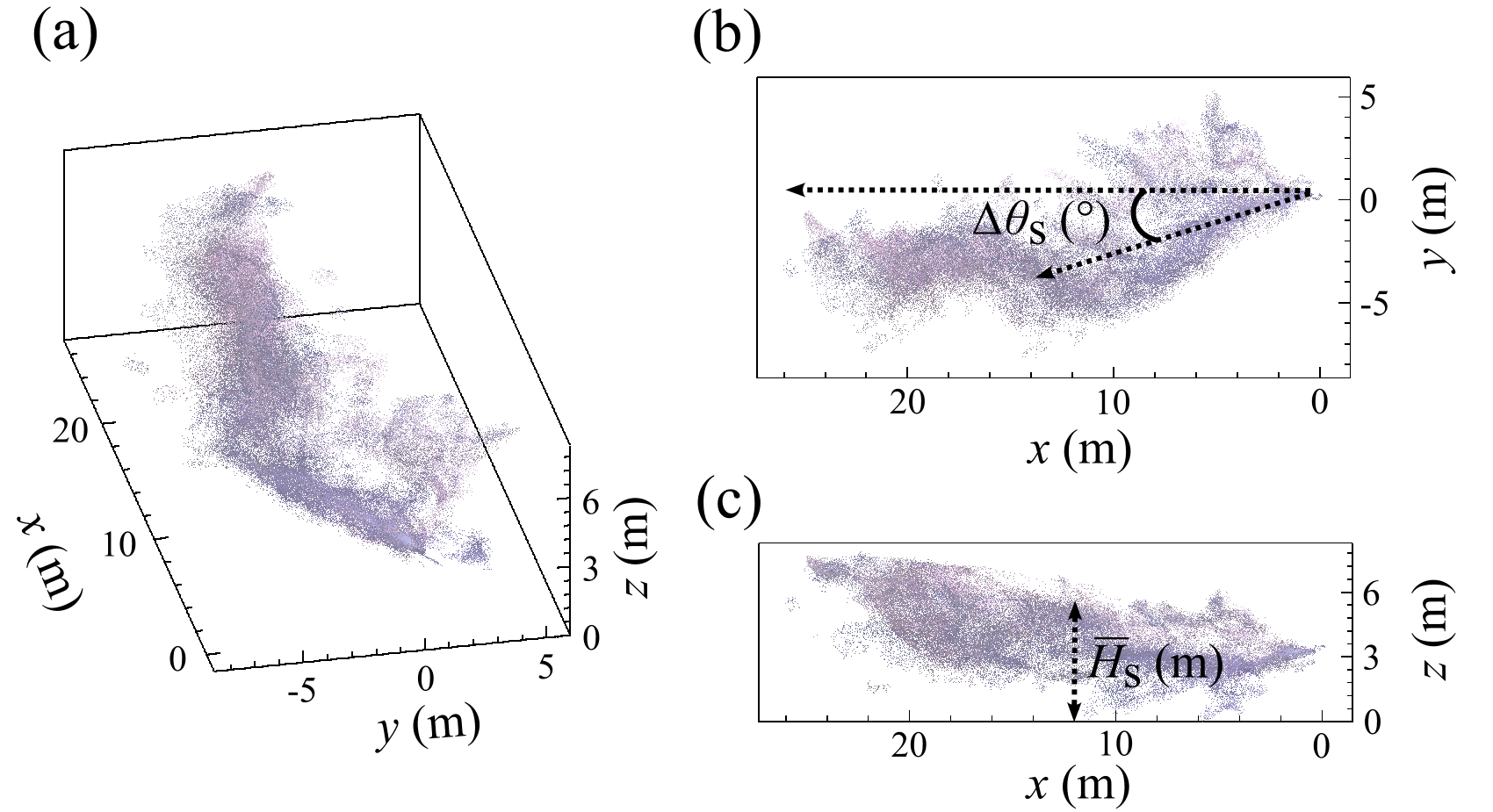}
\caption{Quantitative analysis of reconstructed plume dynamics: (a) Volume trends over time, showing cyclic behavior, (b) Side view (x-z plane) illustrating variations in average plume height, and (c) Top view (x-y plane) depicting angle of deviation and directional changes.}
\label{figure-7}
\end{figure}
\egroup
This section highlights the capability of our drone swarm-based 3D reconstruction system to quantitatively analyze essential plume parameters for controlled burns. From the reconstructed 3D models, we extracted critical metrics: total plume volume $V_{\mathrm{s}}$, angle of deviation (AOD) $\Delta \theta_{\mathrm{s}}$, and average plume height $\overline{H}_{\mathrm{s}}$. Researchers modeling plumes have shown significant interest in studying changes in elevation and volume to better understand plume lifecycles \citep{cao2021simulating, raznjevic2023high}. AOD has been particularly critical in the development of tools like QUIC-Fire, as it captures the influence of wind on particle transport and fire behavior \citep{robinson2023quic}. Guided by these findings, we incorporated the extraction of these parameters into our 3D reconstruction models to enhance the analysis of smoke dynamics. These metrics provide valuable insights into plume behavior, including growth, transport direction, and lofting, which are crucial for applications such as prescribed burn management and forest fire research.

\bgroup
\begin{figure*}[!htbp]
\centering
\includegraphics{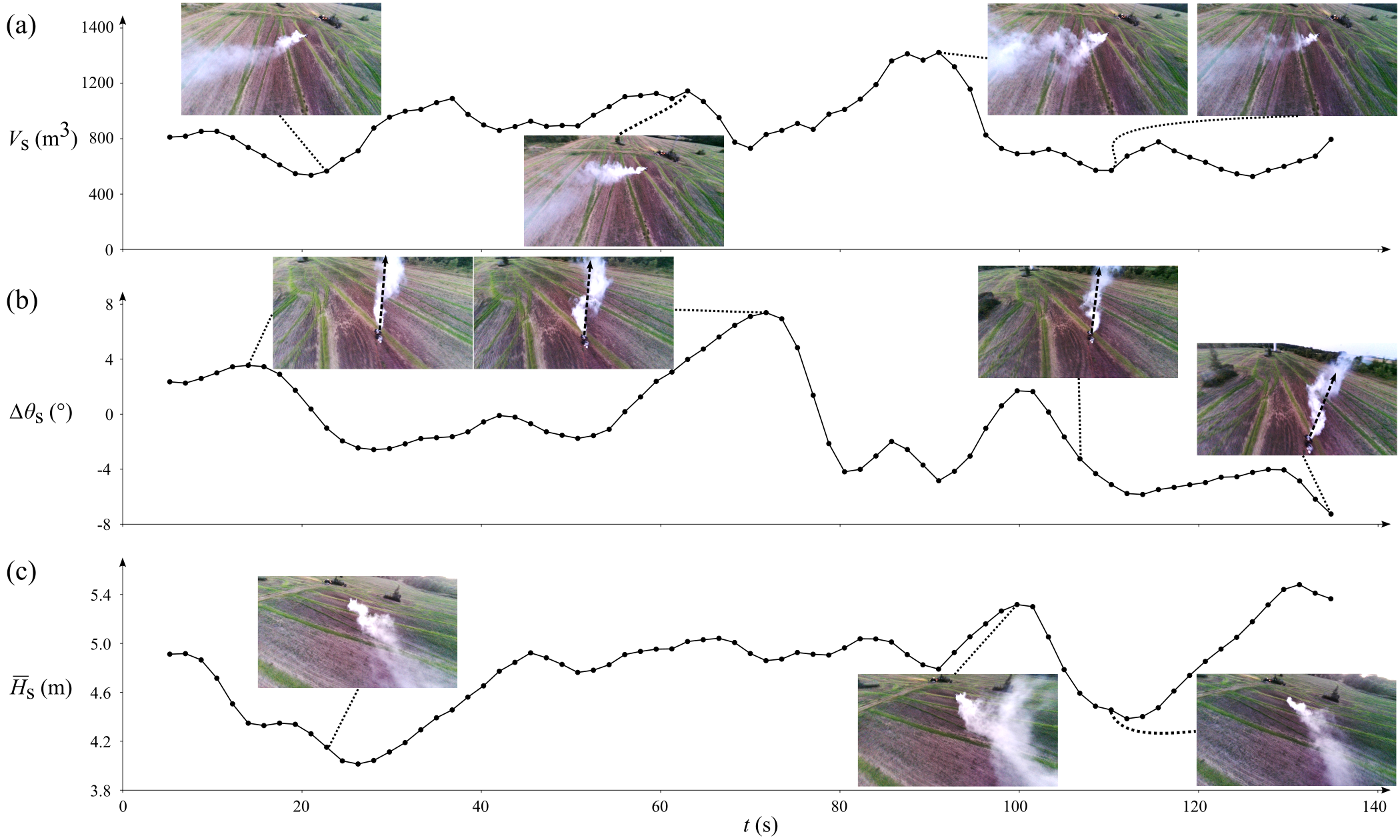}
\caption{Temporal trends in plume characteristics derived from 3D reconstructions, showing the variation in (a) volume ($V_{\mathrm{s}}$), (b) angle of deviation ($\Delta \theta_{\mathrm{s}}$), and (c) average height ($\overline{H}_{\mathrm{s}}$), validated with visual data from individual drone recordings.}
\label{figure-8}
\end{figure*}
\egroup
To calculate these critical parameters of plume dynamics, specific methodologies were applied to the reconstructed 3D data, as illustrated in \textcolor{blue}{Figure~\ref{figure-7}}. The $V_{\mathrm{s}}$ was estimated using the Convex Hull approach, which encloses the plume's data points within the smallest convex shape, providing a practical, though approximate, measure of its spatial boundaries. This method allowed us to plot the plume's volume changes over time. The $\Delta \theta_{\mathrm{s}}$, reflecting the influence of wind on plume direction, was determined by projecting the plume onto a horizontal plane and calculating the average x and y coordinates. A vector connecting these coordinates to the plume's origin was used to compute the angle between this vector and a reference line parallel to the x-axis, representing the plume's average direction. Lastly, the $\overline{H}_{\mathrm{s}}$ of the plume was analyzed by calculating the mean elevation of all points in the cloud and plotting it against time.

To ensure these computed parameters were scaled to real-world dimensions, we utilized the known diameter of the drone trajectories. The NeRF model reconstructs data based on trajectory and pose estimations derived from COLMAP, which adheres to a unified coordinate system. By applying the known real-world diameter of the drone paths as a scaling factor, the reconstructed data could be converted into real-world measurement units, enabling physical interpretation of the plume volume, AOD, and height.

Based on the calculation methods described above, each parameter ($V_{\mathrm{s}}$, $\Delta \theta_{\mathrm{s}}$, $\overline{H}_{\mathrm{s}}$) was computed, and the results are presented in \textcolor{blue}{Figure~\ref{figure-8}}. The trends are analyzed as follows:

1) Volume Change Analysis: As shown in \textcolor{blue}{Figure~\ref{figure-8}a}, the plume's volume exhibits a cyclic pattern over time, with distinct peaks and troughs corresponding to periods of smoke generation and diminishment. These fluctuations align with the operation of the smoke machines, where active emissions produce large, dense plumes, and intervals of reduced output result in smaller, more dispersed plumes. This cyclic behavior reflects the temporal dynamics of the plume, driven by the smoke machine's operational cycles. Validation using drone-captured images confirms this pattern, showcasing high-volume plumes during active phases and diminished plumes during quieter intervals. The alignment between these visual observations and the plotted data supports the accuracy of the extracted volume measurements.

2) AOD Analysis: The angle of deviation, as shown in \textcolor{blue}{Figure~\ref{figure-8}b}, captures directional shifts in the plume's trajectory under the influence of wind. During steady wind conditions, deviations are minimal, with a typical range of \ensuremath{\pm}10\ensuremath{^\circ}. However, when the second smoke machine activates, the plume intensity increases, resulting in higher velocity and longer travel distances. In these cases, the wind's influence becomes more pronounced, causing distinct peaks in the AOD plot. Visual validation highlights this behavior, with the plume initially remaining straight due to the machine's propulsion, then displaying a turning effect as dispersion increases. The visuals also include a reference line that clearly illustrates deviations from the plume's average direction, emphasizing the transition from machine-driven to wind-driven behavior.

3) Average Height Analysis: The trends in average height, depicted in \textcolor{blue}{Figure~\ref{figure-8}c}, closely follow the volume pattern in \textcolor{blue}{Figure~\ref{figure-8}a} for most of the recording period. During active smoke generation, the plume achieves higher elevations, especially when wind direction aligns with the plume's flow. Conversely, reduced emissions result in lower plume heights, with the smoke dispersing rapidly at the far end due to wind effects. Notable exceptions occur when $\overline{H}_{\mathrm{s}}$ remains relatively high despite low $V_{\mathrm{s}}$, which is attributed to narrower plume widths maintaining lofting while reducing overall volume. Validation with drone-captured visuals further supports these observations, illustrating the interplay between smoke generation, vertical expansion, and wind-driven dissipation.

Overall, the deployment of the swarm-based 3D reconstruction system effectively captured and characterized the dynamic nature of smoke plumes, highlighting critical behaviors such as cyclic volume variations, wind-driven directional changes, and the intricate interplay between smoke generation and lofting. Specifically, AOD measurements offered insights into how quickly plumes bend under various atmospheric conditions, and time-resolved 3D reconstructions of plume lofting revealed how far and how quickly smoke can rise under influences like wind shear, terrain, and the smoke source’s output rate \citep{achtemeier2012}. These set of information are especially useful for validating simulation tools such as QUIC-Fire and FIRETEC to ensure the simulated behaviour is matching with the actual variations in terrain changes and wind parameters.

\subsection{Conclusion and Discussion}This study presented a novel drone swarm system for 3D reconstruction of dynamic smoke plumes, combining multi-view imaging with NeRF to achieve high-resolution temporal and spatial plume characterization. The system comprises one manager drone and four worker drones working in a coordinated fashion, with each drone equipped with high-resolution cameras, RTK-enabled GPS for precise positioning, and onboard processing units.Field deployment was conducted to validate both the model's accuracy and the effectiveness of the 3D reconstruction process. To assess reconstruction precision, the system was first tested against a static reference object before applying it to dynamic plume reconstruction. The validation confirmed an average reconstruction error of approximately 1.18\%, demonstrating the system's ability to accurately model 3D structures, ensuring confidence in its application for smoke plume analysis. Following this validation, the system was deployed in capturing dynamic plume characteristics such as cyclic volume variations, wind-driven directional shifts, and the interplay between smoke generation and lofting. The system reconstructed 77 time segments over a two minute and 20 s interval with a temporal resolution of 1.75 s, yielding detailed quantitative data on plume volume, angle of deviation, and average height. These results validate its precision in analyzing highly dynamic and complex plume dispersal patterns.

The multi-view drone swarm imaging system introduced in our study has significant implications for fire management and environmental monitoring. Specifically, by providing high-resolution, time-resolved 3D reconstructions, the system generates unique field data to enhance predictive models such as QUIC-Fire and FIRETEC for prescribed burns and wildfire control, addressing critical gaps in existing fire and smoke simulation tools\citep{gomez2018,kochanski2018experimental, prichard2019fire, gallagher2021, blanco2024}. In addition, the system enables real-time tracking of plume dispersion and air quality monitoring associated with emissions from natural disasters such as wildfires and volcanic eruptions, as well as from controlled burns and industrial processes \citep{butwin2019effects,cao2021simulating}. This capability supports more effective risk assessment, regulatory compliance, and mitigation strategies for air pollution and fire hazards. Beyond its core functionality, the UAV platform offers the flexibility to integrate additional sensing modalities, such as LiDAR and acoustic sensors, further extending its capabilities for multi-modal environmental assessment. These enhancements allow the system to operate effectively in low-visibility conditions, such as dense smoke or nighttime monitoring, and broaden its applicability to complex atmospheric studies and disaster response scenarios. Furthermore, the system is designed to be highly cost-effective. With each drone costing approximately \$1,000 USD, a five-drone swarm presents a significantly more affordable alternative to high-resolution LiDAR or multispectral imaging systems. Even with upgrades for extended flight operations, the overall cost remains substantially lower than traditional approaches, making this UAV swarm system a practical and scalable solution for both real-time hazard response and long-term environmental monitoring.

While this system has demonstrated strong performance, there remains ample opportunity to enhance its adaptability. In particular, its use of fixed circular flight paths can be expanded into waypoint-based or adaptive routes, improving responsiveness to complex or rapidly evolving plume shapes \unskip~\citep{tankasala2022smooth}. Such flexible trajectories would not only refine spatial and temporal coverage—especially in dense regions of a plume—but also help the drones stay aligned with sudden shifts caused by variable winds, all while preserving accurate scaling and georeferencing.

A second avenue for improvement involves camera-pose estimation. Although COLMAP delivers robust results, it can be computationally intensive. Moving toward onboard sensor–based pose estimation would expedite data processing and move the system closer to real-time operation. Similarly, while NeRF-based reconstruction offers high fidelity, it often lacks the speed needed for immediate feedback. Implementing faster variants (e.g., Dynamic NeRF or Instant NGP) could significantly improve efficiency \unskip~\citep{pumarola2021d, muller2022instant}. Moreover, the system’s reliance on visual data alone poses a challenge in low-contrast smoke conditions, where the scarcity of identifiable features can hinder reconstruction \unskip~\citep{mildenhall2021nerf}. Future work will thus explore adaptive exposure control and optimized neural architectures to extend applicability across various plume scenarios.

Looking ahead,  enhancing autonomy, computational efficiency, and near-real-time 3D analysis will further expand the system’s capabilities beyond the applications demonstrated in this study. By improving adaptive navigation, optimizing reconstruction speed, and refining visual processing under challenging conditions, the UAV swarm can become a more versatile tool for large-scale fire management and environmental monitoring. These advancements will enable more effective tracking of emissions from wildfires, volcanic eruptions, prescribed burns, and industrial processes, providing high-resolution air-quality data and actionable insights for hazard response and mitigation.

\section*{Acknowledgements}This work was supported by the NSF Major Research Instrumentation program (NSF-MRI-2018658), which provided crucial funding for the development and deployment of the autonomous drone system for smoke plume characterization. We would also like to extend our gratitude to Sujeendra Ramesh, Sri Ganesh Lalit Aditya Divakarla, and Rammesh Adhav Saravanan  for their invaluable  assistance during the initial phase of the work.
    
\section*{Declaration of generative AI and AI-assisted technologies in the writing process}
During the preparation of this work, the author(s) used ChatGPT for assisting in refining content in this manuscript. After using this tool, the author(s) reviewed and edited the content as needed and take full responsibility for the content of the publication.

\bibliographystyle{elsarticle-harv} 
\bibliography{references.bib} 

\end{document}